\documentclass{article}

\usepackage{microtype}
\usepackage{graphicx}
\usepackage{subfigure}
\usepackage{booktabs} 
\usepackage{color}
\usepackage{framed,multirow}
\usepackage{amssymb}
\usepackage{latexsym}
\usepackage{url}
\usepackage{xcolor}
\usepackage{algorithm}
\usepackage{algorithmic}
\usepackage{enumitem}
\usepackage{amsmath}
\usepackage{amsthm}
\usepackage{makecell}
\usepackage{etoolbox}
\usepackage{tabularx}
\usepackage{array}
\usepackage{float}
\usepackage{multirow}

\usepackage{hyperref}

\usepackage[accepted]{icml2020Arxiv}

\icmltitlerunning{Improve SGD Training via Aligning Mini-batches}

\begin{document}

\twocolumn[
\icmltitle{Improve SGD Training via Aligning Mini-batches}




\begin{icmlauthorlist}
\icmlauthor{Xiangrui Li}{to}
\icmlauthor{Deng Pan}{to}
\icmlauthor{Xin Li}{to}
\icmlauthor{Dongxiao Zhu}{to}
\end{icmlauthorlist}

\icmlaffiliation{to}{Computer Science, Wayne State University}

\icmlcorrespondingauthor{Dongxiao Zhu}{dzhu@wayne.edu}
\icmlkeywords{Machine Learning, ICML}

\vskip 0.3in
]



\printAffiliationsAndNotice{} 

\begin{abstract}
Deep neural networks (DNNs) for supervised learning can be viewed as a pipeline of a feature extractor (i.e. last hidden layer) and a linear classifier (i.e. output layer) that is trained jointly with stochastic gradient descent (SGD). In each iteration of SGD, a mini-batch from the training data is sampled and the true gradient of the loss function is estimated as the noisy gradient calculated on this mini-batch. From the feature learning perspective, the feature extractor should be updated to learn meaningful features with respect to the entire data, and reduce the accommodation to noise in the mini-batch. With this motivation, we propose In-Training Distribution Matching (ITDM) to improve DNN training and reduce overfitting. Specifically, along with the loss function, ITDM regularizes the feature extractor by matching the moments of distributions of different mini-batches in each iteration of SGD, which is fulfilled by minimizing the maximum mean discrepancy. As such, ITDM does not assume any explicit parametric form of data distribution in the latent feature space. Extensive experiments are conducted to demonstrate the effectiveness of our proposed strategy.
\end{abstract}

\section{Introduction}\label{sec:intro}
Recently, deep neural networks (DNNs) have achieved remarkable performance improvements in a wide range of challenging tasks in computer vision \cite{krizhevsky2012alexnet,he2016resnet,huang2019densenet}, natural language processing \cite{sutskever2014sequence,chorowski2015attention} and healthcare informatics \cite{miotto2018deep}. Modern architectures of DNNs usually have an extremely large number of model parameters, which often outnumbers the available training data. Recent studies in theoretical deep learning have shown that DNNs can achieve good generalization even with the over-parameterization \cite{neyshabur2017exploring,olson2018modern}. Although over-parameterization may not be very damaging to DNN's overall generalizability, DNNs can still overfit the noise within the training data (e.g. sampling noise in data collection) due to its highly expressive power. This makes DNNs sensitive to small perturbations in testing data, for example, adversarial samples \cite{goodfellow2014explaining}. To alleviate overfitting of DNNs, many methods have been proposed. These include classic ones such as early stopping, $L_1$ and $L_2$ regularization \cite{goodfellow2016deep}, and more recent ones such as dropout \cite{srivastava2014dropout}, batch normalization \cite{ioffe2015batch} and data-augmentation types of regularization (e.g. cutout \cite{devries2017cutout}, shake-shake \cite{gastaldi2017shake}). There are also other machine learning regimes that can achieve regularization effect such as transfer learning \cite{pan2009transferlearning} and multi-task learning \cite{caruana1997multitask, ruder2017mtl}.

For supervised learning, DNNs can be viewed as a feature extractor followed by a linear classifier on the latent feature space, which is jointly trained using stochastic gradient descent (SGD). When DNNs overfit the training data and a large gap between training and testing loss (e.g. cross-entropy loss for classification) is observed, from the feature learning perspective, it implies mismatching of latent feature distributions between the training and testing data extracted by the feature extractor. Regularization methods mentioned above can reduce such mismatching and hence improve DNNs performance, as the linear classifier can accommodate itself to the latent features to achieve good performance.

In this paper, we propose a different regularization method, called In-Training Distribution Matching (ITDM), that specifically aims at reducing the fitting of noise for feature extraction during SGD training. The idea behind ITDM is motivated by a simple interpretation of the mini-batch update (in addition to the approximation perspective). 

Specifically, in each iteration of SGD, a mini-batch of $m$ samples $\{(x_i,y_i)\}_{i=1}^m$ is sampled from the training data $\{(x_i,y_i)\}_{i=1}^n (n>m)$. The gradient of loss function $L(x,\theta)$ is calculated on the mini-batch, and network parameter $\theta$ is updated via one step of gradient descent (learning rate $\alpha$):
\begin{equation}\label{eq:SGDupdate}
\begin{split}
&\frac{1}{n}\sum_{i=1}^{n}\nabla_\theta L(x_i,\theta) \approx \frac{1}{m}\sum_{i=1}^{m}\nabla_\theta L(x_i,\theta),\\
&\theta \leftarrow \theta - \alpha \cdot\frac{1}{m}\sum_{i=1}^{m}\nabla_\theta L(x_i,\theta).
\end{split}
\end{equation} 
This update (Eq.(\ref{eq:SGDupdate})) can be interpreted from two perspectives. (1) From the conventional approximation perspective, the true gradient of the loss function (i.e. gradient on the entire training data) is approximated by the mini-batch gradient. As each mini-batch contains useful information for the learning tasks and its gradient computation is cheap, large DNNs can be efficiently and effectively trained with modern computing infrastructures. Moreover, theoretical studies in deep learning have shown that the noisiness in the estimated gradient using the randomly sampled mini-batch plays a crucial role in DNNs generalizability \cite{ge2015escaping,daneshmand2018escaping}.
(2) \textbf{Eq. (\ref{eq:SGDupdate}) can also be interpreted as an exact gradient descent update on the mini-batch}. In other words, SGD updates network parameter $\theta$ to achieve maximum improvement in fitting the mini-batch. As each mini-batch is noisy, such exact update inevitably introduces the undesirable mini-batch-dependent noise. In terms of feature learning, the DNN feature extractor can encode the mini-batch noise into the feature representations. 

These two perspectives enable us to decompose the SGD update intuitively as:
\[\text{Training improvement}+\text{(possible) Mini-batch overftting}.\]
A natural question then to ask is \textit{``Can we reduce the mini-batch overfitting?"} to reduce the mini-batch dependence in SGD update Eq. (\ref{eq:SGDupdate}). One solution to this problem is batch normalization (BN) \cite{ioffe2015batch}. In their seminal paper, the internal covariate shift is observed due to the distribution shift of the activation of each hidden layer from mini-batch to mini-batch. Under our decomposition, this phenomenon is closely related to the mini-batch overfitting as networks have to adjust parameter $\theta$ to fit the mini-batches. To reduce the distribution shift, \cite{ioffe2015batch} introduces the BN layer that fixes the means and variances of activations of hidden layers.

Different from BN, the proposed ITDM directly reduces the mini-batch overfitting by matching its latent feature distribution with another mini-batch. In this paper, we only consider the feature representation from the last hidden layer. Ideally, if the distribution $P(h)$ of latent feature $h$ is known as a prior, we could explicitly match the mini-batch feature $h_{\text{mb}}$ with $P(h)$ via maximum likelihood. However, in practice, $P(h)$ is not known or does not even have an analytic form. To tackle this problem, we utilize the maximum mean discrepancy (MMD) \cite{gretton2012kernel} from statistical hypothesis testing for the two-sample problem. Our motivation of using MMD as the matching criterion is: \textit{if the SGD update using one mini-batch A is helpful for DNNs learning good feature representations with respect to the entire data, then for another mini-batch B, the mismatch of latent feature distributions between A and B should not be significant}. In this way, we can reduce mini-batch overfitting by forcing accommodation of SGD update to B and reducing dependence of the network on A. In terms of model training, MMD has two advantages: (1) it enables us to avoid the presumption for $P(h)$ and (2) the learning objective of MMD is differentiable which can be jointly trained with $L(x,\theta)$ by backpropagation. Note that ITDM is not a replacement of BN. In fact, ITDM can benefit from BN when BN helps improving the feature learning for DNNs.

We summarize our contributions as follows. (1) We propose a training strategy ITDM for training DNNs. ITDM augments conventional SGD with regularization effect by additionally forcing feature matching of different mini-batches to reduce mini-batch overfitting. ITDM can be combined with existing regularization approaches and applied on a broad range of network architectures and loss functions. (2) We conduct extensive experiments to evaluate ITDM. Results on different benchmark datasets demonstrate that training with ITDM can significantly improve DNN performances, compared with conventional training strategy (i.e. perform SGD only on the loss function).

\section{Related Work}
In this section, we first review regularization methods in deep learning. Our work utilizes MMD and hence is also related to the topic of distribution matching, so we also review its related works that is widely studied under the context of domain adaption and generative modeling. 

With limited amount of training data, training DNNs with a large number of parameters usually requires regularization to reduce overfitting. Those regularization methods include class ones such as $L_1/L_2$-norm penalties and early stopping \cite{hastie2009elements,goodfellow2016deep}. For deep learning, many new approaches are proposed motivated by the SGD training dynamics. For example, dropout \cite{srivastava2014dropout} and its variants \cite{gao2019demystifying, ghiasi2018dropblock} achieves regularization effect by reducing the co-adaption of hidden neurons of DNNs. In the training process, dropout randomly sets some hidden neurons' activation to zero, resulting in an averaging effect of a number of sub-networks. \cite{ioffe2015batch} proposes batch normalization (BN) to reduce the internal covariate shift caused by SGD. By maintaining the mean and variance of mini-batches, BN regularizes DNNs by discouraging the adaption to the mini-batches. Label smoothing \cite{szegedy2016ls} is another regularization technique that discourage DNNs' over-confident predictions for training data. Our proposed ITDM is partially motivated by the covariate shift observed by \cite{ioffe2015batch}. ITDM achieves regularization by reducing DNN's accommodation to each mini-batch in SGD as it is an exact update for that mini-batch.

To match the distribution of different mini-batches, ITDM uses MMD as its learning objective. MMD \cite{gretton2007kernel, gretton2012kernel} is a probability metric for testing whether two finite sets of samples are generated from the same distribution. With the kernel trick, minimizing MMD encourages to match all moments of the data empirical distributions. MMD has been widely applied in many machine learning tasks. For example, \cite{li2015gmmd} and \cite{li2017mmdgan} use MMD to train unsupervised generative models by matching the generated distribution with the data distribution. Another application of MMD is for the domain adaption. To learn domain-invariant feature representations, \cite{long2015dan} uses MMD to explicitly match feature representations from different domains. Our goal is different from those applications. In ITDM, we do not seek exact distribution matching. Instead, we use MMD as a regularization to improve SGD training. 

\section{In-Training Distribution Matching}
In this section, we first provide an introduction of maximum mean discrepancy for the two-sample problem from the statistical hypothesis testing. Then we present our proposed ITDM for training DNNs using SGD, along with some details in implementation. 

\begin{figure*}[t]
	\centering
	\includegraphics[scale=0.4]{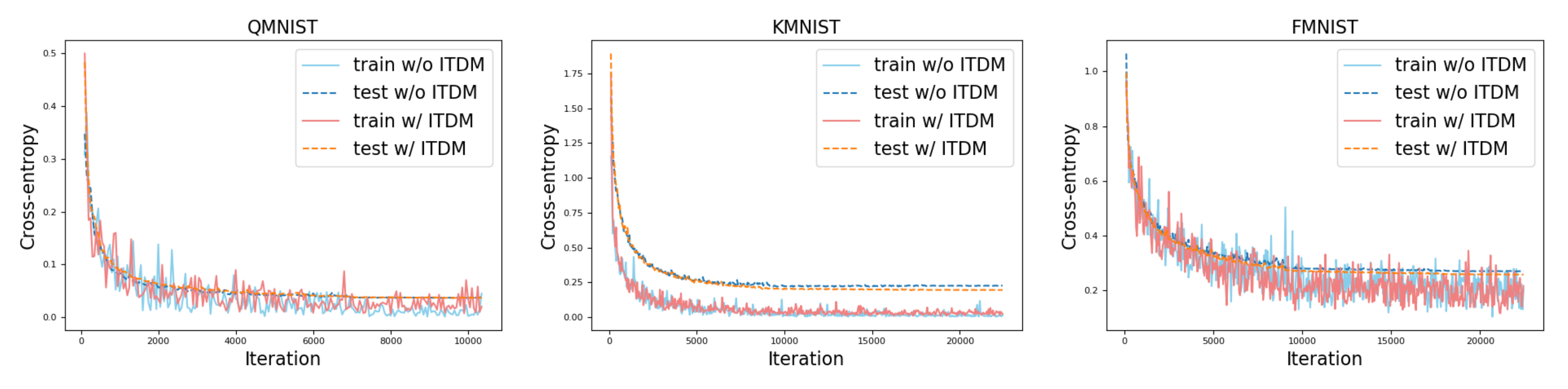}
	\caption{Training and testing cross-entropy for QMNIST, KMNIST and FMNIST.}
	\label{fig:PrimaryCurve}
\end{figure*}

\begin{figure*}[h!]
	\centering
	\includegraphics[scale=0.4]{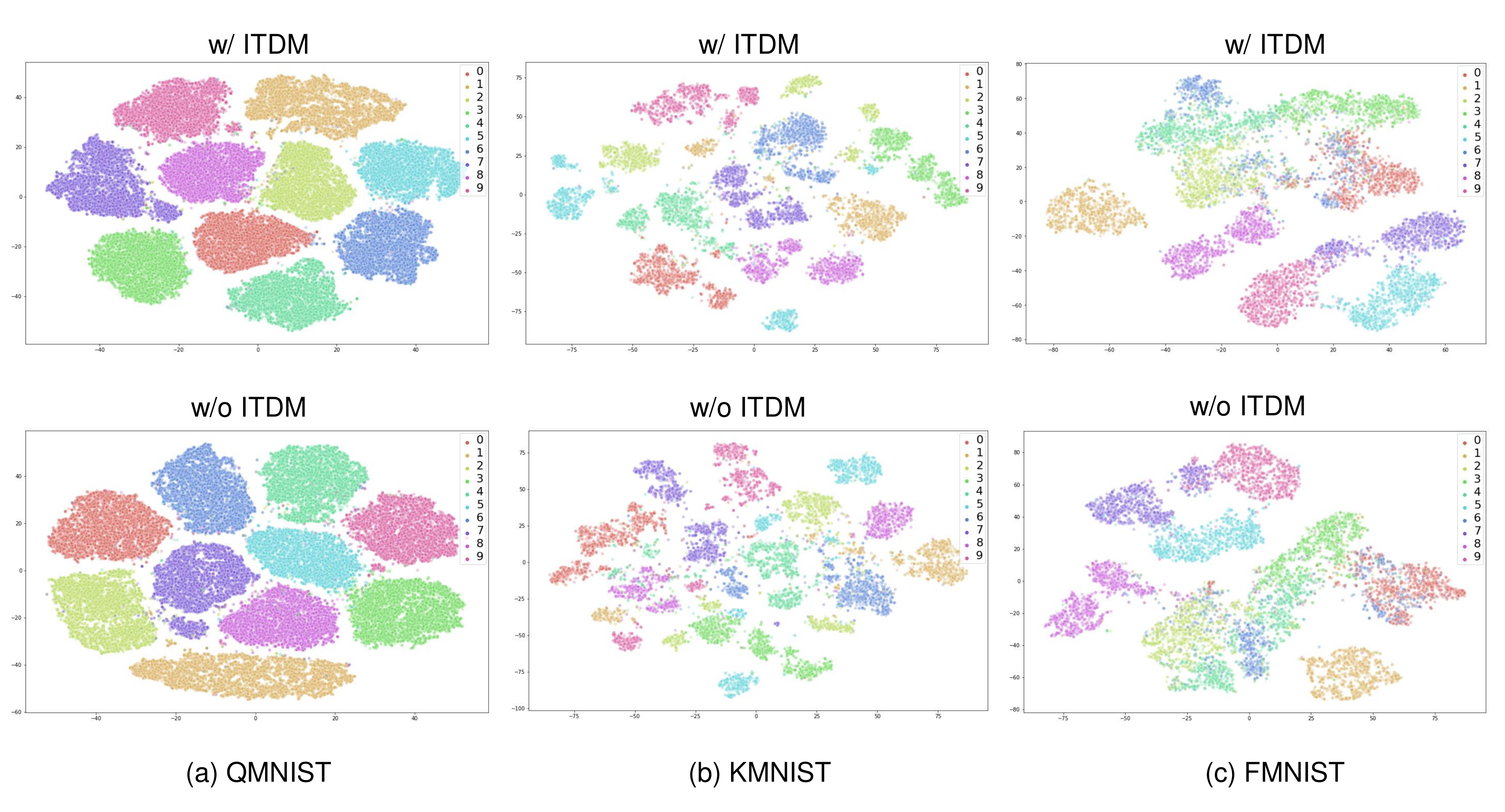}
	\caption{T-SNE \cite{maaten2008tsne} plot for training CNN with and without ITDM-j using cross-entropy as loss function on the standard testing data of QMNIST, KMNIST and FMNIST. Compared with the conventional training that only performs SGD on the loss function, ITDM has slightly clearer margin and boundaries in the visualization.}
	\label{fig:PrimaryTSNE}
\end{figure*}

\subsection{Maximum Mean Discrepancy}
Given two finite sets of samples $S_1 = \{x_i\}_{i=1}^n$ and $S_2 = \{y_i\}_{i=1}^m$, MMD \cite{gretton2007kernel,gretton2012kernel} is constructed to test whether $S_1$ and $S_2$ are generated from the same distribution. MMD compares the sample statistics between $S_1$ and $S_2$, and if the discrepancy is small, $S_1$ and $S_2$ are then likely to follow the same distribution.

Using the kernel trick, the empirical estimate of MMD \cite{gretton2007kernel} w.r.t $S_1$ and $S_2$ can be rewritten as:
\begin{equation*}
\begin{split}
\text{MMD}(S_1, S_2) &= \big[\frac{1}{n^2}\sum_{i,j=1}^{n}k(x_i,x_j) + \frac{1}{m^2}\sum_{i,j=1}^{m} k(y_i,y_j)\\
                     & - \frac{2}{mn}\sum_{i=1}^{n}\sum_{j=1}^{m}k(x_i,y_j) \big]^{1/2},
\end{split}
\end{equation*}
where $k(\cdot,\cdot)$ is a kernel function. \cite{gretton2007kernel} shows that if $k$ is a characteristic kernel, then asymptotically MMD$=0$ if and only $S_1$ and $S_2$ are generated from the same distribution. A typical choice of $k$ is the Gaussian kernel with bandwidth parameter $\sigma$:
\[k(x,y) = \exp(-\frac{||x-y||^2}{\sigma}).\]
The computational cost of MMD is $O((m+n)^2)$. In ITDM, this is not problematic as typically only a small number of samples in each mini-batch (e.g. 100)  is used in SGD.

\subsection{Proposed ITDM}
The idea of ITDM, as explained in Section \ref{sec:intro}, is to reduce the DNN adaption to each mini-batch if we view the SGD iteration as an exact update for that mini-batch. In terms of feature learning, we attempt to train the feature extractor to encode less mini-batch dependent noise into the feature representation. From the distribution point of view, the latent feature distribution of the mini-batch should approximately match with, or more loosely, should not deviate much from that of the entire data. However, matching with the entire data has some disadvantages. If MMD is used as matching criterion and training data size (say $n$) is large, the time complexity for MMD is not desirable (i.e. $O(n^2)$). For computational efficiency, an analytic form of the latent feature distribution can be assumed but we will be at the risk of misspecification. As such, we propose to use a different mini-batch only for latent feature matching (and not for classification loss function). As seen in the experiments, this strategy can significantly improve the performance in terms of loss values on the independent testing data.

More formally, let $f_\theta(x)$ be a convolutional neural network model for classification that is parameterized by $\theta$. It consists of a feature extractor $h = E_{\theta_e}(x)$ and a linear classifier $C_{\theta_c}(h)$ parameterized by $\theta_e$ and $\theta_c$ respectively. Namely, $f_\theta(x)= C_{\theta_c}(E_{\theta_e}(x))$ and $\theta = \{\theta_e, \theta_c\}$. Without ambiguity, we drop $\theta$ in $f, E$ and $C$ for notational simplicity. 

In each iteration of SGD, let $S_1 = \{(x_i^1,y_i^1)\}_{i=1}^{m_1}$ be the mini-batch of $m_1$ samples. Then the loss function using cross-entropy (CE) on $S_1$ can be written as 
\begin{equation}\label{eq:cross-entropy}
L_{mb}(\theta) = -\frac{1}{m_1}\sum_{i=1}^{m_1} \log f_{y_i^1}(x_i^1),
\end{equation}
where $f_{y_i^1}(x_i^1)$ is the predicted probability for $x_i^1$'s true label $y_i^1$. SGD performs one gradient descent step on $L_{mb}$ w.r.t $\theta$ using Eq. (\ref{eq:SGDupdate}).

To reduce $\theta$'s dependence on $S_1$ in this exact gradient descent update, we sample from the training data another mini-batch $S_2= \{(x_i^2,y_i^2)\}_{i=1}^{m_2}$ to match the latent feature distribution between $S_1$ and $S_2$ using MMD:
\begin{equation} \label{eq:matchlossJ}
\begin{split}
&H_1 = \{h_i^1=E(x_i^1):i=1,\cdots,m_1\},\\
&H_2 = \{h_i^2=E(x_i^2):i=1,\cdots,m_2\},\\
&\text{Match}_j(\theta_e; H_1, H_2)  = \text{MMD}(H_1,H_2).
\end{split}
\end{equation}

Our proposed ITDM modifies the conventional gradient descent step in SGD by augmenting the cross-entropy loss (Eq. (\ref{eq:cross-entropy})) with the matching loss, which justifies the name of ITDM:
\begin{equation}\label{eq:itdmj}
\theta \leftarrow \theta - \alpha \nabla_\theta \big[ L_{mb}(\theta) + \lambda\text{Match}_j(\theta_e;H_1, H_2) \big],
\end{equation}
where $\lambda$ is the tuning parameter controlling the contribution of Match$_j$. We call the update in Eq. (\ref{eq:itdmj}) as \textbf{ITDM-j} since we match the \textbf{j}oint distribution of $(x,y)$ without differentiating each class in the mini-batches $S_1$ and $S_2$. Note that in ITDM-j, mini-batch $S_2$ is not used in the calculation of cross-entropy loss $L_{mb}(\theta)$.

\begin{table}[h!]
	\caption{Classification accuracy (in \%) and CE loss trained with and without ITDM-j, on the testing data of QMNIST, KMNIST and FMNIST. Results are reported as the average of last 10 iterations.}
	\label{tb:primaryResults}
	\begin{center}
		\begin{small}
			\begin{sc}
				\begin{tabular}{lccccr}
					\toprule
					&    & QMNIST & KMNIST & FMNIST \\
					\midrule
				 \multirow{2}{*}{Acc} &	w/ ITDM   & 98.94 & 94.42  & 90.79 \\
					                  & w/o ITDM  & 98.86 & 94.19  & 90.52 \\
				    \midrule
				 \multirow{2}{*}{CE}  & w/ ITDM   & 0.037  & 0.196  & 0.257 \\
				                      & w/o ITDM  & 0.037  & 0.227  & 0.269 \\
					\bottomrule
				\end{tabular}
			\end{sc}
		\end{small}
	\end{center}
\end{table} 

\textbf{Initial results using ITDM-j}  To test the effectiveness of ITDM, we performed initial experiments using three MNIST-type datasets: QMNIST \cite{yadav2019cold}, KMNIST \cite{clanuwat2018deep} and FMNIST \cite{xiao2017fmnist}. We trained a simple CNN of two convolutional layers with and without ITDM under the exactly same setting. Experiment details are provided in the supplemental material. Table \ref{tb:primaryResults} shows the classification accuracy and cross-entropy loss (i.e. negative log-likelihood) on the standard testing data. Figure \ref{fig:PrimaryCurve} plots the curve of cross-entropy loss for both training and testing data. It can be seen that ITDM-j achieves better results compared with the conventional training. Notably, ITDM-j has overall smaller loss on the testing data which implies the model makes correct predictions with larger probability. To verify this, we plot the T-SNE embedding of latent features in Figure \ref{fig:PrimaryTSNE}. We see that with ITDM, the latent feature for each class has slightly clearer margin and boundaries. This implies that ITDM can help SGD converge to a better local minimum.

\textbf{Class-conditional ITDM} For classification tasks, we could utilize the label information and further refine the match loss as a sum of class-conditional match loss, termed as \textbf{ITDM-c} (using the notation in Eq. (\ref{eq:matchlossJ})):
\begin{equation}\label{eq:matchlossC}
\begin{split}
&H_1^k = \{h_i^1: y_i=k, i=1,\cdots,m_1\} (k=1,\cdots,K),\\
&H_2^k = \{h_i^2: y_i=k, i=1,\cdots,m_2\} (k=1,\cdots,K),\\
&\text{Match}_c (\theta_e; H_1, H_2) = \frac{1}{K}\sum_{k=1}^K \text{Match}_j(\theta_e; H_1^k, H_2^k),
\end{split}
\end{equation}
where $K$ is the total number of classes and $y_i=k$ the true label of sample $x_i$. The ITDM-c update in SGD is 
\begin{equation}\label{eq:itdmc}
\theta \leftarrow \theta - \alpha \nabla_\theta \big[ L_{mb}(\theta) + \lambda\text{Match}_c(\theta_e;H_1, H_2) \big].
\end{equation}
The overall training procedure of ITDM is summarized in Algorithm \ref{alg:ITDM}.

Compared with ITDM-j, ITDM-c has two advantages. (1) With the implicitly utilization of label information of mini-batch $S_2$, ITDM-c can help DNN learn better feature representation by focusing on the in-class distribution matching. (2) With in-class matching, the computational cost for calculating MMD is reduced from $O((m_1+m_2)^2)$ to $O(Km_H^2)$ where $m_H = \max_k \{\#H_i^k:i=1,2, k=1,\cdots,K\}$. 

\begin{algorithm}[t]
	\caption{IN-Training Distribution Matching}
	\label{alg:ITDM}
	\begin{algorithmic}[1]
		\STATE {\bfseries Input:} training data $\{(x_i, y_i)\}$
		\STATE {\bfseries Initialization:} model parameter $\theta$
		\FOR{each epoch}
		\FOR{each mini-batch $S_1$}
		\STATE Sample another mini-batch $S_2$
		\STATE Calculate $L_{mb}$ using $S_1$ (Eq. (\ref{eq:cross-entropy}))
		\STATE Calculate $H_1$ and $H_2$ for $S_1$ and $S_2$ (Eq. (\ref{eq:matchlossJ}))
		\STATE Calculate Match$_j$ or Match$_c$ (Eq. (\ref{eq:matchlossJ}) or (\ref{eq:matchlossC}))
		\STATE Perform ITDM update using Eq. (\ref{eq:itdmj}) or (\ref{eq:itdmc})
		\ENDFOR
		\ENDFOR
	\end{algorithmic}
\end{algorithm}

\subsection{Implementation Considerations}
\textbf{Bandwidth parameter $\sigma$ in Gaussian kernel} The performance MMD as a metric of matching two samples is sensitive to the choice of the bandwidth parameter $\sigma$ when Gaussian kernel is used. Since we generally do not have the prior knowledge about the latent feature distribution, we follow the practice in \cite{gretton2007kernel,gretton2012kernel,long2015dan} that takes the heuristic of setting $\sigma$ as the median squared distance between two samples. In ITDM, $\sigma$ is not prefixed but rather estimated in each iteration of SGD w.r.t two mini-batches. 

We check the gradient of Gaussian kernel $k(x,y)$ to justify this choice of $\sigma$:
\begin{equation}\label{eq:kernelgradient}
\nabla_x k(x,y) = -2k(x,y)\frac{x-y}{\sigma}.
\end{equation}
For a fixed $\sigma$, if $x$ and $y$ are either close to or far from each other, $\nabla_xk(x,y)$ is small and hence provides little information in the backpropagation. By setting $\sigma$ as the running median squared distance between random mini-batches, the MMD loss can automatically adapt itself and maintain useful gradient information.

It is worth mentioning that MMD with Gaussian kernel may not effectively carry gradient information for hard samples, as the latter are usually close to the decision boundary and far away from the majority of samples. The reason is due to small $\nabla_x k(x,y)$ if $x$ and $y$ is far from each other. To remedy this, we use a mixture of $g$ Gaussian kernels with different ranges of $\sigma$s:
\[k_{\text{mix}}(x,y) = \frac{1}{g}\sum_{i=1}^{g}k_{\sigma_i}(x,y).\]


\textbf{Mini-batch size} When used as the training objective for distribution matching, MMD usually requires large batch-size for effective learning. For example, \cite{li2015gmmd} sets the batch size to 1000. However, our goal of ITDM is not for exact distribution matching, but rather as a regularization to reduce the mini-batch overfitting in SGD update. In our experiments, we set the batch size following the common practice (e.g 150) and it works well in practice without introducing many computational burdens.

\section{Experiments}
In this section, we evaluate the ITDM strategy and compare its performance with the vanilla SGD training on several benchmark datasets of image classification, i.e. \textbf{w/ ITDM \textit{v.s} w/o ITDM}. Specifically, ITDM-c is tested as it provides implicit label information with better supervision in the training process. We implement our codes in Pytorch \cite{paszke2019pytorch} and utilize Nivdia RTX 2080TI GPU for computation acceleration.

\subsection{Datasets}
We test ITDM on four benchmark datasets Kuzushiji-MNIST (KMNIST) \cite{clanuwat2018deep}, Fashion-MNIST (FMNIST) \cite{xiao2017fmnist}, CIFAR10 \cite{krizhevsky2009cifar} and STL10 \cite{coates2011stl10}. KMNIST and FMNIST are two gray-scale image datasets that are intended as alternatives to MNIST. Both datasets consist of 70000 (28 $\times$ 28) images from 10 different classes of Japanese character and clothing respectively, among which 60000 are used for training data and the remaining 10000 for testing data. CIFAR10 is a colored image dataset of 32 $\times$ 32 resolution. It consists of 50000 training and 10000 testing images from 10 classes. STL10 is another colored image dataset where each image is of size 96 $\times$ 96. Original STL10 has 100000 unlabeled images, 5000 labeled for training and 8000 labeled for testing. In our experiment, we only use the labeled subset for evaluation.

\subsection{Implementation Details}
Through all experiments, the optimization algorithm is the standard stochastic gradient descent with momentum and the loss function is cross-entropy (CE) loss. In ITDM-c, CE loss is further combined with the matching loss (Eq. (\ref{eq:itdmc}) in each iteration. 

On KMNIST and FMNIST, we build a 5-layer convolutional neural network (CNN) with batch normalization applied. Detailed architecture is provided in the supplemental material.  Momentum is set to 0.5, batch size 150, number of epochs 50, initial learning rate 0.01 and multiplied by 0.2 at 20th and 40th epoch. No data augmentation is applied. For CIFAR10 and STL10, we use publicly available implementation of VGG13 \cite{simonyan2014very}, Resnet18 \cite{he2016resnet} and Mobilenet \cite{howard2017mobilenet}. All models are trained with 150 epochs, SGD momentum is set to 0.5, initial learning rate is 0.5 and multiplied by 0.1 every 50 epochs, batch size 150. We resize STL10 to 32 $\times$ 32. For colored image datasets, we use random crop and horizontal flip for data augmentation.

In all experiments, networks are trained with vanilla SGD and ITDM SGD under the exactly same setting (learning rate, batch size et al.). For the tuning parameter $\lambda$ in ITDM, we test \{0.2, 0.4, 0.6, 0.8, 1\} for checking ITDM's sensitivity to it. Note that when $\lambda=0$, ITDM is equivalent to vanilla SGD training. For the bandwidth parameter in the mixture of Gaussian kernels, we use 5 kernel $k_{\sigma_i}$ with $\{\sigma_i = 2^i\sigma_{\text{Med}}: i=0,\cdots,4\}$. We utilize the standard train/test split given in the benchmark datasets and train all models once on the training data. Performances are evaluated on the testing data and reported as the average of the last 10 iterations.

\subsection{Results\protect\footnote{Complete results are provided in supplemental materials.}}

For predictive performance in the table, we report the best (B) and worst (W) Top-1 accuracy trained with ITDM, and their corresponding $\lambda$ and cross-entropy (CE) loss values, which is equivalent to negative log-likelihood. For better comparison, the performance difference of $\Delta$ between with and without ITDM is also reported with $\uparrow$ indicating significant improvement from ITDM and $\downarrow$ otherwise.

\begin{table}[t]
	\caption{Accuracy (in \%, larger is better)  and CE (smaller is better) on KMNIST testing data.}
	\label{tb:KMNISTResults}
	\begin{center}
		\begin{small}
			\begin{sc}
				\begin{tabular}{lc|cc|cc}
					\toprule
					& $\lambda$ &  Acc    & $\Delta$ & CE & $\Delta$ \\
					\midrule
					w/o ITDM   & -         &  95.57  &     -    & 0.183    &    - \\
					w/ ITDM (B) & 0.8       &  95.79  &  0.22$\uparrow$     & 0.170    &   0.013$\uparrow$  \\
					w/ ITDM (W) & 0.4       &  95.59  &  0.02   & 0.162    &   0.031$\uparrow$  \\
					\bottomrule
				\end{tabular}
			\end{sc}
		\end{small}
	\end{center}
\end{table} 

\begin{figure}[t]
	\centering
	\includegraphics[width=\columnwidth]{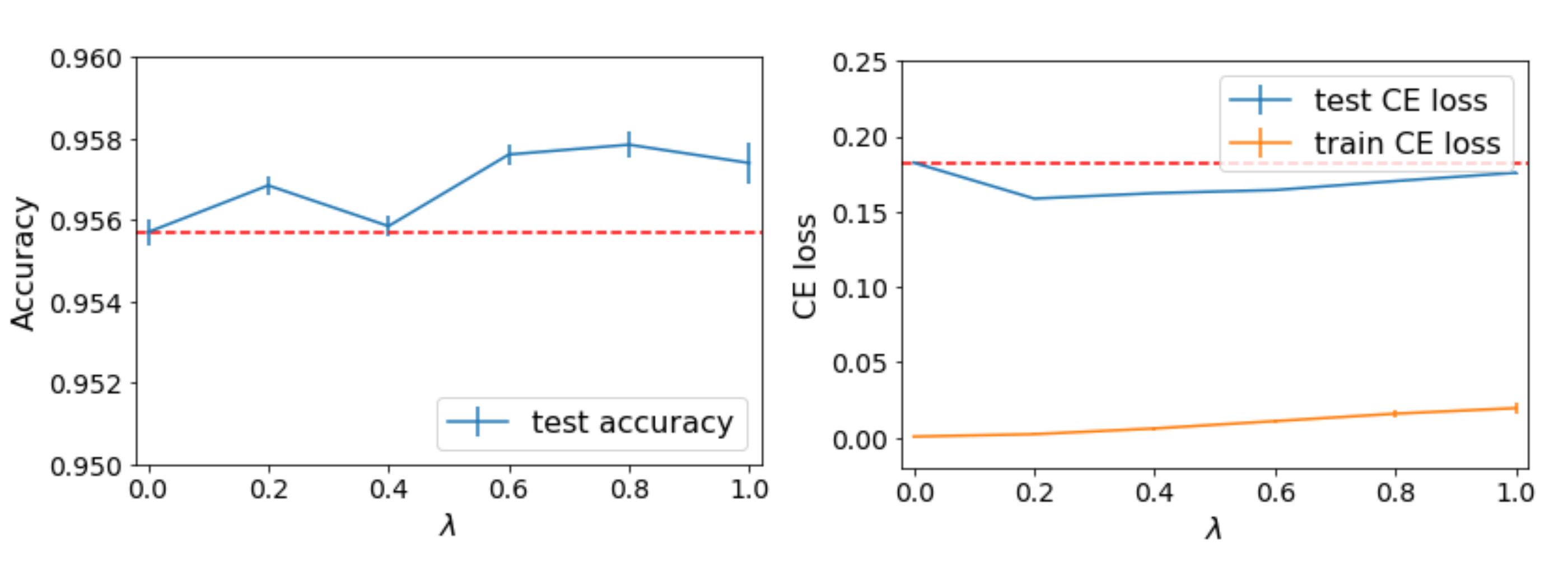}
	\caption{Accuracy and CE loss (with one standard deviation error bar) w.r.t different $\lambda$ values on KMNIST. $\lambda=0$ is equivalent to the vanilla SGD training.}
	\label{fig:KMNISTLambdaCurve}
\end{figure}

\textbf{KMNIST} Table \ref{tb:KMNISTResults} shows the predictive performance for KMNIST. From the table, we see that training with ITDM achieves better results in terms of accuracy and CE. Even in the worst case, ITDM has comparable accuracy with that of the vanilla SGD training. In Figure \ref{fig:KMNISTLambdaCurve}, we also plot the accuracy, training and testing loss (after optimization converges) against $\lambda$. From the figure, we have the following observations. (1) On KMNIST, training with ITDM is not very sensitive to $\lambda$, which at least has comparable performance in terms of accuracy, and always has smaller CE loss. As CE is equivalent to negative log-likelihood, smaller CE value implies the network makes predictions on testing data with higher confidence on average. (2) As $\lambda$ increases, the training CE loss also increases. This is expected as in each iteration of ITDM, there is a tradeoff between the CE and match loss. Since a larger CE implies smaller likelihood, ITDM has a regularization effect by alleviating the over-confident predictions on training data.

\begin{figure}[t]
	\centering
	\includegraphics[width=\columnwidth]{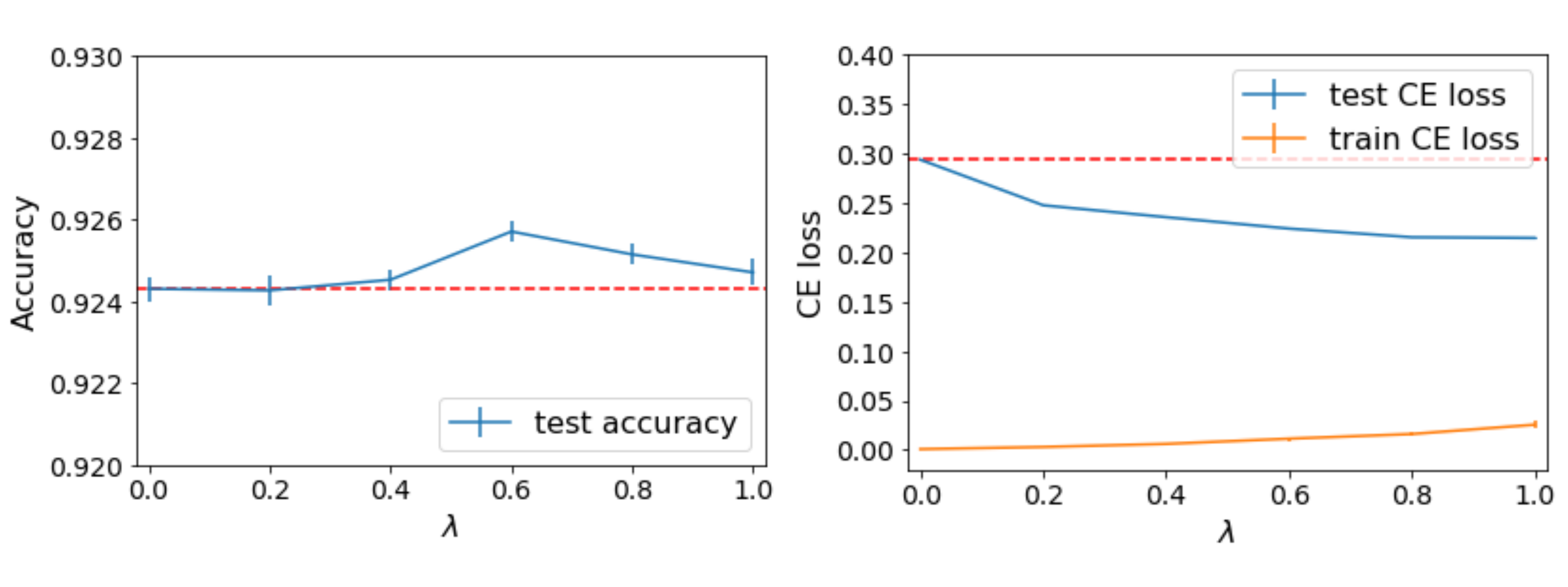}
	\caption{Accuracy and CE (with error bar) w.r.t different $\lambda$ values on FMNIST. $\lambda=0$ is equivalent to the vanilla SGD training.}
	\label{fig:FMNISTLambdaCurve}
\end{figure}

\textbf{FMNIST} Table \ref{tb:FMNISTResults} shows the predictive performance for FMNIST. We plot in Figure \ref{fig:FMNISTLambdaCurve} the accuracy, training and testing loss. As can be seen from the table and figure, ITDM generally does not damage the predictive accuracy. Similar to KMNIST, ITDM always has smaller CE values. However it does not necessarily lead to accuracy gain. The possible reason is that, FMNIST has a significant number of hard samples (e.g. those from pullover, coat and shirt classes). Though ITDM can always lead to prediction with stronger confidence, it still misses those hard samples as MMD may not be able to effectively capture their information in the training process (Eq. (\ref{eq:kernelgradient})).

\begin{table}[t]
	\caption{Accuracy (in \%, larger is better)  and CE loss (smaller is better) on FMNIST testing data.}
	\label{tb:FMNISTResults}
	\begin{center}
		\begin{small}
			\begin{sc}
				\begin{tabular}{lc|cc|cc}
					\toprule
					& $\lambda$ &  Acc    & $\Delta$ & CE & $\Delta$ \\
					\midrule
					w/o ITDM   & -         &  92.43  &     -    & 0.294    &    - \\
					w/ ITDM (B) & 0.6       &  92.57  &  0.14$\uparrow$     & 0.224    &   0.070$\uparrow$  \\
					w/ ITDM (W) & 0.2       &  92.42  &  0.01    & 0.248    &   0.046$\uparrow$  \\
					\bottomrule
				\end{tabular}
			\end{sc}
		\end{small}
	\end{center}
\end{table} 

\begin{figure}[t]
	\centering
	\includegraphics[width=\columnwidth]{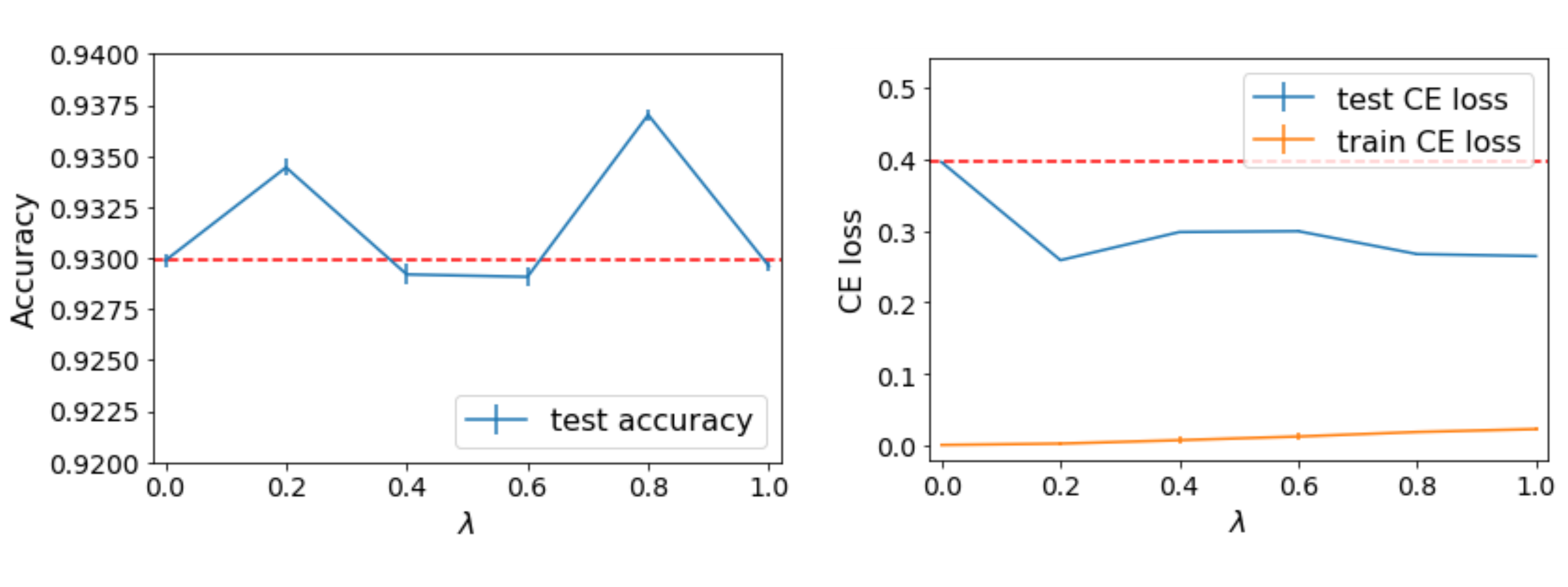}
	\caption{Resnet18 accuracy and CE (with error bar) w.r.t different $\lambda$ values on CIFAR10.}
	\label{fig:CIFAR10LambdaCurve}
\end{figure}

\textbf{CIFAR10} In Table \ref{tb:CIFARResults}, we present the performance of Resnet18, VGG13 and Mobilenet on CIFAR10. For Resnet and Mobilenet, the overall performances of training with and without ITDM in terms of accuracy are comparable across all $\lambda$ values. In particular, when $\lambda$ is set with a relatively large value of 0.8 or 1, ITDM can further improve the accuracy by a margin 0.71\% for Resnet and 0.82\% for Mobilenet. For VGG13, training with ITDM gives higher accuracy and worse when $\lambda=0.8$ and $\lambda=0.2$ respectively. We also plot the CE loss for different $\lambda$s in Figure \ref{fig:CIFAR10LambdaCurve} w.r.t Resnet (for illustration purpose). Comparing with vanilla SGD training, we see that training with ITDM results in significant gain in CE, regardless of network architecture: Resent $32.6\% (0.129/0.396)$, VGG $29.4\%$ and Mobilenet $30.6\%$. This pattern also holds even if ITDM does not outperform vanilla SGD training in terms of accuracy: Resnet $24.5\%(0.097/0.396)$, VGG $25.8\%$ and Mobilenet $17.6\%$. On the other hand, as $\lambda$ increases, the training loss also increases. A closer gap between training and testing loss usually implies better generalization as it means a closer distribution match between train and testing data. From this perspective, ITDM can regularize DNNs to learn better feature representations with better generalizability. 

\begin{table}[t]
	\caption{Accuracy (in \%, larger is better)  and CE loss (smaller is better) of Resnet18, VGG13 and Mobilenet on CIFAR10.}
	\label{tb:CIFARResults}
	\begin{center}
		\begin{small}
			\begin{sc}
				\begin{tabular}{lc|cc|cc}
					\toprule
					\multicolumn{6}{c}{Resnet18} \\
					\midrule
					& $\lambda$ &  Acc    & $\Delta$ & CE & $\Delta$ \\
					\midrule
					w/o ITDM   & -         &  92.99  &     -    & 0.396    &    - \\
					w/ ITDM (B) & 0.8       &  93.70  &  0.71$\uparrow$     & 0.267    &   0.129$\uparrow$  \\
					w/ ITDM (W) & 0.6       &  92.91  &  0.08$\downarrow$   & 0.299    &   0.097$\uparrow$  \\
					\midrule
					
					\midrule
					\multicolumn{6}{c}{VGG13} \\
					\midrule
					& $\lambda$ &  Acc    & $\Delta$ & CE & $\Delta$ \\
					\midrule
					w/o ITDM   & -         &  92.49  &     -    & 0.473    &    - \\
					w/ ITDM (B) & 0.8       &  92.72  &  0.23$\uparrow$     & 0.334    &   0.139$\uparrow$  \\
					w/ ITDM (W) & 0.2       &  92.34  &  0.15$\downarrow$    & 0.351    &   0.122$\uparrow$  \\
					\midrule
					
					\midrule
					\multicolumn{6}{c}{Mobilenet} \\
					\midrule
					& $\lambda$ &  Acc    & $\Delta$ & CE & $\Delta$ \\
					\midrule
					w/o ITDM   & -         &  88.55  &     -    & 0.615    &    - \\
					w/ ITDM (B) & 1.0       &  89.37 &  0.82$\uparrow$     & 0.427    &   0.188$\uparrow$  \\
					w/ ITDM (W) & 0.2       &  88.76  &  0.21$\uparrow$    & 0.507   &   0.108$\uparrow$  \\
					\bottomrule
				\end{tabular}
			\end{sc}
		\end{small}
	\end{center}
\end{table} 

\begin{figure*}[t]
	\centering
	\includegraphics[scale=0.7]{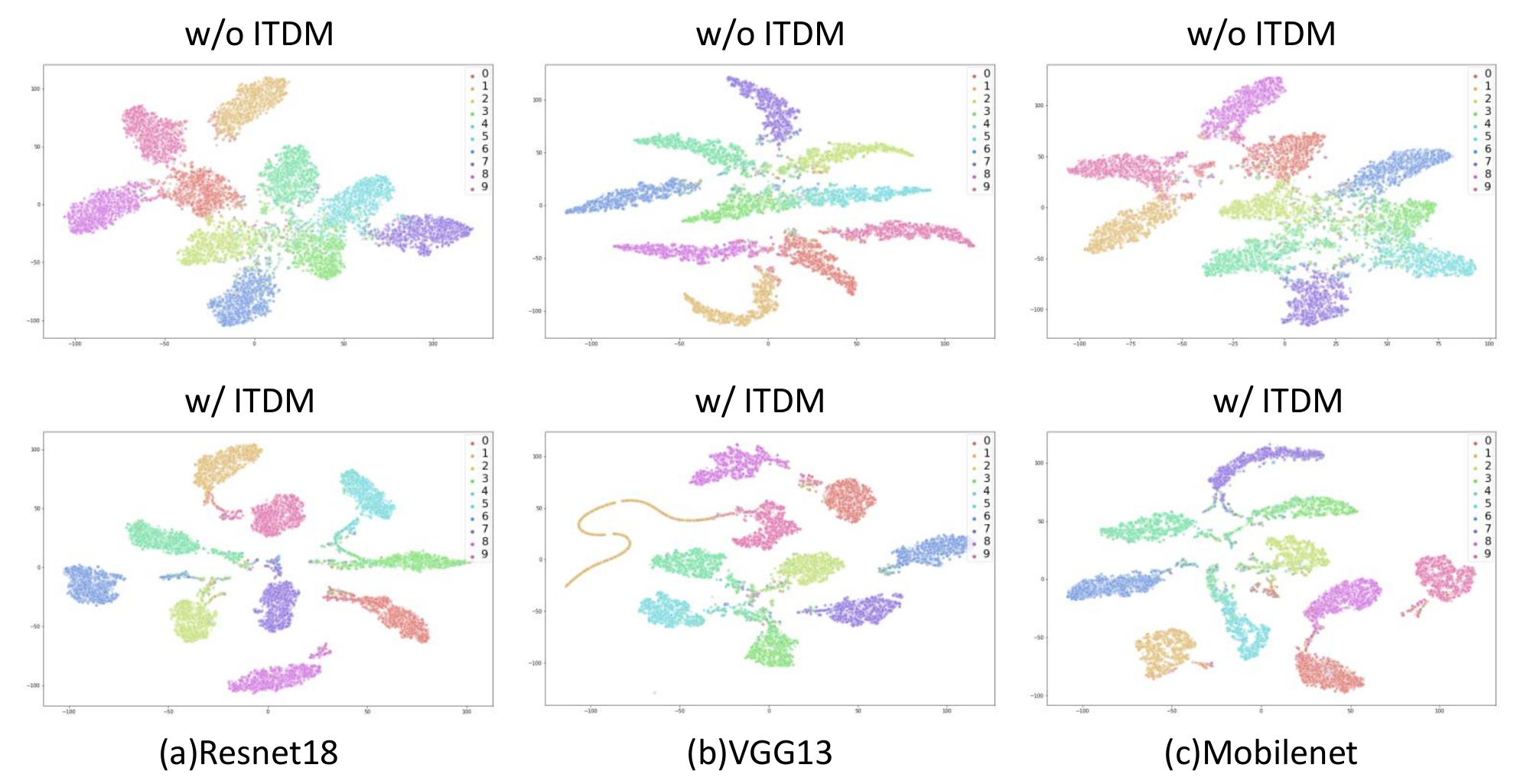}
	\caption{T-SNE plot for CIFAR10 testing data. Networks are trained with $\lambda$ that achieves best accuracy in Table \ref{tb:CIFARResults}.}
	\label{fig:CIFARTSNE}
\end{figure*}
\begin{figure}[t]
	\centering
	\includegraphics[width=\columnwidth]{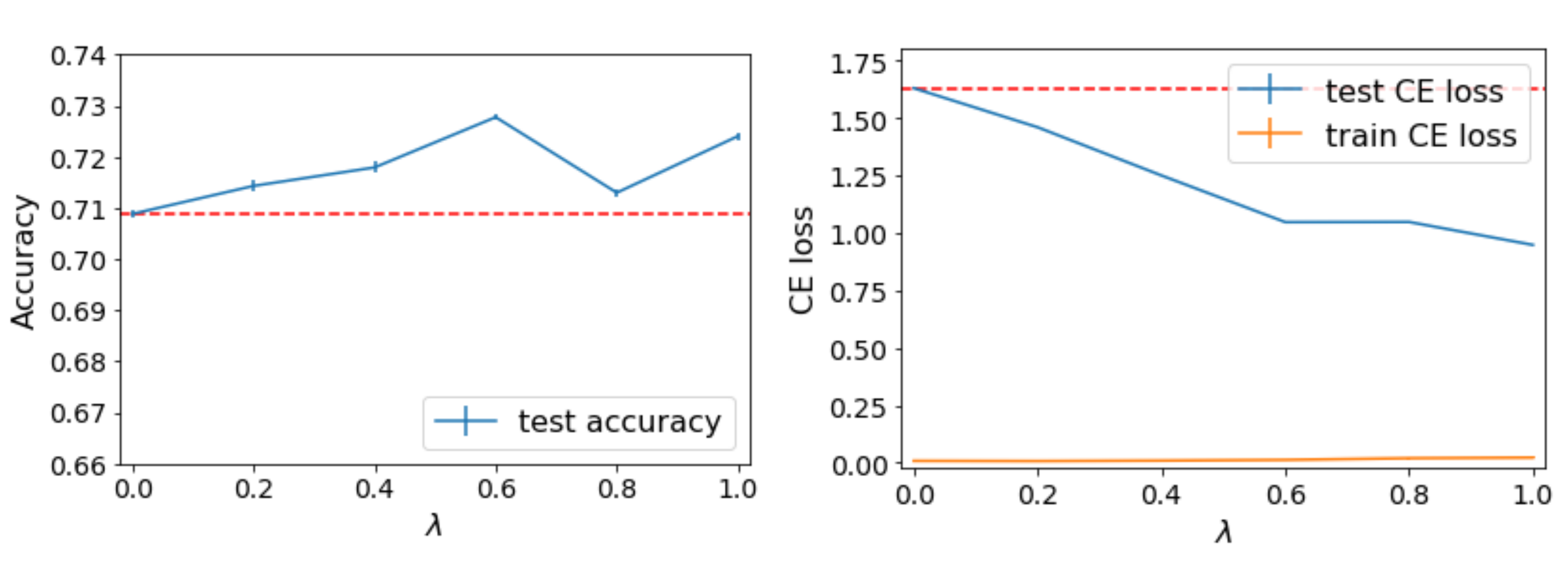}
	\caption{Resnet18 accuracy and CE loss (with error bar) w.r.t different $\lambda$ values on STL10.}
	\label{fig:STL10LambdaCurve}
\end{figure}

\textbf{STL10} The results on STL10 is shown in Table \ref{tb:STLResults}. Similar to CIFAR10, a larger value of $\lambda$ results in higher accuracy with significant margin for ITDM, i.e., Resnet $1.9\%$, VGG13 $1.4\%$ and Mobilenet $2.93\%$, whereas a smaller value leads to performance drop, i.e., Resnet $2.47\%$ and Mobilent $1.23\%$. In terms of CE loss, ITDM always outperforms vanilla SGD training, i.e., Resnet $35.6\% (0.581/1.630)$, VGG $39.5\%$ and Mobilenet $20.6\%$ (Figure \ref{fig:STL10LambdaCurve}).

\begin{table}[t]
	\caption{Accuracy (in \%, larger is better)  and CE loss (smaller is better) of Resnet18, VGG13 and Mobilenet on STL10 testing data.}
	\label{tb:STLResults}
	\begin{center}
		\begin{small}
			\begin{sc}
				\begin{tabular}{lc|cc|cc}
					\toprule
					\multicolumn{6}{c}{Resnet18} \\
					\midrule
					& $\lambda$ &  Acc    & $\Delta$ & CE & $\Delta$ \\
					\midrule
					w/o ITDM   & -         &  70.88  &     -    & 1.630    &    - \\
					w/ ITDM (B) & 0.6       &  72.78  &  1.90$\uparrow$     & 1.049    &   0.581$\uparrow$  \\
					w/ ITDM (W) & 0.8       &  71.29  &  0.41$\uparrow$   & 1.048    &   0.582 $\uparrow$ \\
					\midrule
					
					\midrule
					\multicolumn{6}{c}{VGG13} \\
					\midrule
					& $\lambda$ &  Acc    & $\Delta$ & CE & $\Delta$ \\
					\midrule
					w/o ITDM   & -         &  74.40  &     -    & 1.545    &    - \\
					w/ ITDM (B) & 0.8       &  75.80  &  1.40$\uparrow$     & 0.934    &   0.611$\uparrow$  \\
					w/ ITDM (W) & 0.4       &  74.46  &  0.06    & 1.110    &   0.435$\uparrow$  \\
					\midrule
					
					\midrule
					\multicolumn{6}{c}{Mobilenet} \\
					\midrule
					& $\lambda$ &  Acc    & $\Delta$ & CE & $\Delta$ \\
					\midrule
					w/o ITDM   & -         &  59.09  &     -    & 2.144    &    - \\
					w/ ITDM (B) & 0.6       &  62.02 &  2.93$\uparrow$     & 1.603   &   0.541$\uparrow$  \\
					w/ ITDM (W) & 0.8       &  58.93 &  0.16$\downarrow$    & 1.638   &   0.506$\uparrow$ \\
					\bottomrule
				\end{tabular}
			\end{sc}
		\end{small}
	\end{center}
\end{table} 

\subsection{Analysis}
Through the extensive experimental results across a broad range of datasets, we observe that ITDM with larger $\lambda$ values tends to have better performances when compared with smaller $\lambda$ values, and outperforms the vanilla SGD training. Since  we use ITDM-c in the experiments, a plausible reason for this phenomenon is that ITDM-c provides implicit supervision in the learning process by matching two random, noisy mini-batches from the \textit{same} class. With larger $\lambda$s, ITDM can benefit from the stronger implicit supervision and hence improve network performance.

Another phenomenon is that ITDM can reduce the testing CE loss significantly, in particular for CIFAR10 and STL10 datasets. Given a sample $(x,y=k)$, its CE loss is calculated as $-\log f_k(x)$, where $f_k$ is the predicted probability for $x$'s true class label $k$. A smaller CE value implies a larger probability $f_k$. From the geometric perspective, samples from the same class should stay close and those from different classes are expected to stay far apart in the feature space (so that $f_k$ output by softmax is large). To confirm this, we visualize the distribution of CIFAR10 testing samples with T-SNE \cite{maaten2008tsne} in Figure \ref{fig:CIFARTSNE}. From the figure, ITDM learns feature representation that is much tighter with clearer inter-class margin than that learned by vanilla SGD training. We also can gain some insight on why ITDM achieves impressive improvement in CE loss but not as much in accuracy: For each class, ITDM effectively captures the ``typical pattern" of each class and the majority of samples are hence clustered closely, but ITDM also misses some hard samples that overlap with other classes. Overall, ITDM still outperforms vanilla SGD training and can be used as a promising training prototype that is capable of learning more discriminative features.

\section{Conclusion}
In this paper, we propose a new training strategy with regularization effect, ITDM, as an alternative to vanilla SGD training. ITDM augments vanilla SGD with a matching loss that uses MMD as the objective function. By forcing the matching of two different mini-batches, ITDM reduces the possible mini-batch overfitting in vanilla SGD. Experimental results demonstrate its excellent performance on classification tasks, as well as its impressive feature learning capacity. There are two possible directions for our future studies. The first one is to improve ITDM that can learn hard sample more effectively. The second one is potential ITDM application in learning form poisoned datasets as ITDM tends capture the major pattern in the dataset.

\bibliography{ITDMRef}
\bibliographystyle{icml2020}

\end{document}